# A Contextual Bandit Approach for Stream-Based Active Learning

Linqi Song
Electrical Engineering Department
University of California, Los Angeles, USA
Email: songlinqi@ucla.edu

Jie Xu
Electrical and Computer Engineering Department
University of Miami, USA
Email: jiexu@miami.edu

*Abstract*—Contextual bandit algorithms – a class of multi-armed bandit algorithms that exploit the contextual information – have been shown to be effective in solving sequential decision making problems under uncertainty. A common assumption adopted in the literature is that the realized (ground truth) reward by taking the selected action is observed by the learner at no cost, which, however, is not realistic in many practical scenarios. When observing the ground truth reward is costly, a key challenge for the learner is how to judiciously acquire the ground truth by assessing the benefits and costs in order to balance learning efficiency and learning cost. From the information theoretic perspective, a perhaps even more interesting question is how much efficiency might be lost due to this cost. In this paper, we design a novel contextual bandit-based learning algorithm and endow it with the active learning capability. The key feature of our algorithm is that in addition to sending a query to an annotator for the ground truth, prior information about the ground truth learned by the learner is sent together, thereby reducing the query cost. We prove that by carefully choosing the algorithm parameters, the learning regret of the proposed algorithm achieves the same order as that of conventional contextual bandit algorithms in cost-free scenarios, implying that, surprisingly, cost due to acquiring the ground truth does not increase the learning regret in the long-run. Our analysis shows that prior information about the ground truth plays a critical role in improving the system performance in scenarios where active learning is necessary.

## I. INTRODUCTION

Contextual bandits [1][2][3] is a powerful machine learning framework for modeling and solving a large class of sequential decision making problems under uncertainty, ranging from content recommendation, online advertising, stream mining, to decision support for clinical diagnosis [4] and personalized education [5]. In a typical setting, a task arrives to the system with certain contextual information (e.g. incoming user's age, gender, search and purchase history etc. in online content recommendation), then the system pulls an arm from a possibly very large arm space (e.g. recommend a piece of online content from a large content pool). A reward is later realized depending on the context value and the selected arm. The objective of a learner (or a learning algorithm) is to make arm selection decisions based on the history of context-arm-reward realizations to minimize the learning regret (i.e. the gap of achievable reward compared with certain benchmarks).

A common assumption made in the literature is that the reward of each task is observed by the learner at no cost,

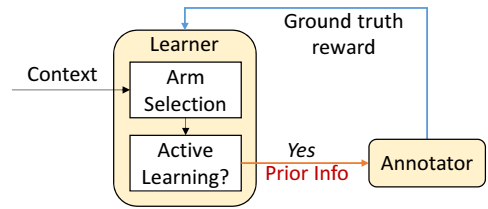

Fig. 1. Sequential decision making with active learning. The learner decides arm selection and whether to acquire the ground truth reward (active learning).

thereby allowing the learner to fully and freely utilize this information. While this assumption holds true in some application scenarios, it hardly captures the reality in many others in which observing the ground truth reward requires substantial manpower, time, energy and/or other resources. For instance, to calculate the reward in stream mining systems, human experts are needed to manually annotate the ground truth labels of the mining tasks. Therefore, in addition to carefully deciding which arm to pull, the learner also has to actively and judiciously acquire the ground truth rewards from an annotator by assessing the benefits and costs of obtaining them in these application scenarios [6][7]. Figure 1 illustrates the considered active learning scenario.

In this paper, we design a learning algorithm, called Contextual Bandits with Active Learning (CB-AL), that accomplishes the aforementioned task. We prove that CB-AL is order-optimal in terms of the learning regret, which matches that of conventional contextual bandits in cost-free scenarios. The key to achieving the optimal regret order by our proposed algorithm is that the query about the ground truth reward is sent to the annotator together with some prior information about this reward. Although the learner does not directly observe the reward realization by selecting an arm, it is learning the distribution of the reward as it learns the optimal arm to pull, and this statistical information can be utilized to reduce the cost of acquiring the ground truth reward by an annotator. This is in stark contrast with conventional active learning literature where the cost of acquiring the ground truth is constant [8][9][10].

Our algorithm is able to effectively deal with large context and arm spaces. To this end, our algorithm divides time into epochs and the context/arm spaces are adaptively partitioned across epochs. The partitions become finer and finer as the epoch grows. Within each epoch, our algorithm first explores various arm clusters (defined for each arm subspace) to learn their reward estimates for each context cluster (defined for each context subspace) and removes suboptimal arm clusters during the course of learning. When the remaining arm clusters are learned to be optimal or near-optimal, the algorithm enters an exploitation phase in which the arm cluster removal operation stops and acquiring the ground truth rewards is no longer needed for the remaining time slots in the current epoch, thereby maximizing the reward and minimizing the query cost concurrently. To optimize the overall long term performance and minimize the long-term learning regret, our algorithm carefully designs control functions that determine what arm clusters are optimal, near-optimal and suboptimal.

The remainder of this paper is organized as follows. Section II formulates the problem and defines the learning regret. Section III describes our algorithm whose regret performance is analyzed in Section IV. Section V provides illustrative numerical results followed by conclusions in Section VI.

## II. PROBLEM FORMULATION

### A. System Model

We consider a discrete time system where time is divided into slots $t = 1, 2, ...$. The arm space is a bounded space $\mathcal{K}$ with covering dimension $d_K$. The context space is a bounded space $\mathcal{X}$ with covering dimension $d_X$. For any context $x \in \mathcal{X}$, the reward of selecting arm $k \in \mathcal{K}$ is $r(x, k) \in [0, 1]$, which is sampled according to some underlying but unknown distribution $f(x, k)$. The expected value of $r(x, k)$ is denoted as $\mu(x, k)$, which is unknown too. We assume that the reward value space is $[0, 1]$ for the ease of exposition but this assumption can be relaxed to account for any bounded interval.

In the **conventional** contextual bandits setting, the following events occur in sequence in each time slot $t$: (1) A context $x_t \in \mathcal{X}$ arrives; (2) An arm $k_t \in \mathcal{K}$ is selected; (3) The (ground-truth) reward $r(x_t, k_t)$ is generated according to $f(x_t, k_t)$ and is **observed** by the learner at **no cost** as feedback, which provides information for future arm selections.

In our considered setting, $r(x_t, k_t)$ is not observed for free. Instead, there is a cost associated with requesting the ground truth reward. Therefore, there is a need to actively and judiciously decide when to request the ground truth reward to balance the learning efficiency and the cost minimization. Thus, in addition to deciding which arm $k_t$ to choose, the learner also has to decide whether or not to query the ground truth reward at a cost, denoted by $q_t \in \{0, 1\}$, where $q_t = 1$ stands for requesting and $q_t = 0$ stands for not requesting.

We consider that the query cost is not fixed, but a function of the prior information about the ground truth, which is updated as learning goes on. The intuition is that if the prior information is more informative, then the query cost should be smaller. In particular, we define the prior information about the reward $r(x_t, k_t)$ as a tuple $(a_t, b_t, \delta_t)$, which represents that the expected reward $\mu(x_t, k_t)$ is in the region $[a_t, b_t]$ with probability at least $1 - \delta_t$. The query cost is then defined as a convex increasing function of the confidence interval $b_t - a_t$ and the significance level $\delta_t$ of the following form:

$$c_t = c[(b_t - a_t)^{\beta_1} + \eta \delta_t^{\beta_2}] \quad (1)$$

where $c > 0$, $\beta_1 \geq 1$, $\beta_2 \geq 1$, $\eta > 0$ are constant parameters. Therefore, a larger confidence interval $b_t - a_t$ and a smaller confidence level $1 - \delta_t$ result in a higher query cost. We choose this form of query cost because it captures the reality to a large extent and also is amenable to our subsequent analysis. Let $\hat{r}^t$ denote the observed reward in time slot $t$, which is $\hat{r}^t = r(x_t, k_t)$ if $q_t = 1$, and $\hat{r}^t = \emptyset$ if $q_t = 0$.

Given the context arrival process, the selected arm sequence and the observed reward sequence, the history by time slot $t$ is defined as

$$h^{t-1} = \{(x_1, k_1, \hat{r}_1), ..., (x_{t-1}, k_{t-1}, \hat{r}_{t-1}))\}, \forall t > 1 \quad (2)$$

and $h^0 = \emptyset$ for $t = 1$. The set of all possible histories is denoted by $\mathcal{H}$. An algorithm $\pi$ is a mapping $\pi : \mathcal{H} \times \mathcal{X} \to \mathcal{K} \times \{0, 1\}$, which selects an arm and decides whether or not to query given the history and the current context. For the ease of exposition, we separately write $\pi_K^t = \pi_K(h^{t-1}, x_t)$ and $\pi_q^t = \pi_q(h^{t-1}, x_t)$ for the arm selection component and the query decision component of the algorithm, respectively.

### B. Learning Regret

We use the total expected payoff (i.e. the reward minus the query cost) to describe the performance of an algorithm $\pi$. The total expected payoff up to time slot $T$ is thus

$$U_\pi(T) = \mathbb{E} \sum_{t=1}^{T} [r(x_t, k_t) - c_t q_t] \quad (3)$$

where the expectation is taken over the context arrival process and the reward distributions. We compare an algorithm with the static-best oracle policy $\pi^*$ which knows the reward distributions *a priori*. Therefore, in each time slot $t$, the oracle policy selects the arm $k_t^* = \arg\max_k \mu(x_t, k_t)$ that maximizes the expected reward. Clearly, since the oracle knows the reward distributions, there is no need for it to query the ground truth to learn about them. Therefore, $q_t^* = 0, \forall t$. The learning regret of an algorithm $\pi$ is defined as

$$R_\pi(T) = U_{\pi^*}(T) - U_\pi(T) \quad (4)$$

As a widely-adopted assumption in contextual bandits literature [1][3], the reward function is assumed to satisfy a Lipschitz condition with respect to both the context and the arm. This assumption is formalized as follows.

**Assumption 1.** *For any two contexts $x, x' \in \mathcal{X}$ and two arms $k, k' \in \mathcal{K}$, the expected rewards satisfy*

$$|\mu(x, k) - \mu(x', k)| \leq L_X \|x - x'\| \quad (5)$$
$$|\mu(x, k) - \mu(x, k')| \leq L_K \|k - k'\| \quad (6)$$

*where $L_X, L_K$ are the Lipschitz constants for the context space and the arm space, respectively.*

## III. THE ALGORITHM

In this section, we describe the proposed contextual bandits algorithm with active learning (CB-AL).

### A. Useful Notions

First, we introduce some useful notions for the algorithm.

**Context/Arm Space Partition**. Time slots are grouped into epochs. The $i$-th epoch lasts for $T_i = 2^i$ time slots. At the beginning of each epoch, the context space and the arm space are partitioned into small subspaces. A context (arm) space is called a context (arm) cluster. The context/arm space partition is kept unchanged throughout the entire epoch. Formally, the partition of the context space for epoch $i$ is denoted by $\mathcal{P}_X(i) = \{\mathcal{X}_1, \mathcal{X}_2, ..., \mathcal{X}_{M_i}\}$ consisting of $M_i$ subspaces. Similarly, the partition of the arm space for epoch $i$ is denoted by $\mathcal{P}_K(i) = \{\mathcal{K}_1, \mathcal{K}_2, ..., \mathcal{K}_{N_i}\}$ consisting of $N_i$ subspaces. The radius of a context cluster $\mathcal{X}_m$ is half of the maximum distance between any two context points in the cluster, i.e.

$$\rho_{\mathcal{X}_m} = 0.5 \sup_{x, x' \in \mathcal{X}_m} \|x - x'\| \quad (7)$$

The radius of an arm cluster is defined similarly. The context/arm space partitioning is performed such that the context/arm clusters satisfy $\forall m = \{1, ..., M_i\}, \rho_{\mathcal{X}_m,i} = T_i^{-\alpha} \triangleq \rho_{X,i}$ and $\forall n = \{1, ..., N_i\}, \rho_{\mathcal{K}_n,i} = T_i^{-\alpha} \triangleq \rho_{K,i}$, where $\alpha \in (0, 1)$.

**Active Arm Cluster**. At the beginning of each epoch, all arm clusters according to the arm space partitioning are set to be active. We denote the set of active arm clusters with respect to context cluster $\mathcal{X}_m$ in epoch $i$ by $\mathcal{A}_m(i)$. The active arm cluster set will be updated as time goes by according to the learning outcome. Some arm clusters will be learned to be suboptimal and hence will be de-activated (i.e. be removed from $\mathcal{X}_m$) and will not be selected by the algorithm in the remaining time slots of the current epoch.

**Round**. A round $s_m(i)$ is defined for each context cluster $\mathcal{X}_m$ in each epoch $i$, which consists a number of $|\mathcal{A}_m(i)|$ time slots. Thus, in each round $s_m(i)$, each active arm cluster in $\mathcal{A}_m(i)$ is selected once. Therefore, even in the same epoch, the length of a round $s_m(i)$ may change due to the updating of the active arm cluster set $\mathcal{A}_m(i)$.

**Control Functions**. There are two important control functions in our algorithm. The first control function, denoted by $D_i(i, s_m(i))$, is used to de-active arm clusters, with respect to each context cluster $\mathcal{X}_m$, that are learned to be suboptimal depending on the epoch index $i$ and the round index $s_m(i)$. In particular, $D_i(i, s_m(i))$ has the form

$$D_1(i, s_m(i)) = \epsilon(i) + [2D(s_m(i)) + 2L_X \rho_{X,i} + 2L_K \rho_{K,i}]$$

where $\epsilon(i) = LT_i^{-\alpha}$ is a small positive value for epoch $i$, and $D(s_m(i)) = \sqrt{\ln(2T_i^{1+\gamma})/2s_m(i)}$. Here $L = L(c) > 4L_X + 4L_K$ and $\gamma \in (0, 1)$ are constants.

The second control function, denoted by $D_2(i, s_m(i))$, is used to determine when to stop querying the ground truth reward. When the stopping condition is satisfied, the algorithm stops de-activating arm clusters and enters a pure exploitation phase for the remaining time slots of the current epoch for the context cluster $\mathcal{X}_m$. In particular, $D_2(i, s_m(i))$ has the form

$$D_2(i, s_m(i)) = 2\epsilon(i) - [2D(s_m(i)) + 2L_X \rho_{X,i} + 2L_K \rho_{K,i}]$$

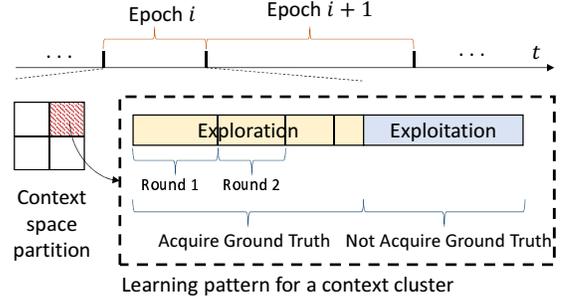

Fig. 2. Contextual Bandits Learning with Active Learning

**Sample Mean Reward**. The sample mean reward of an arm cluster $\mathcal{K}_n$ with respect to a context cluster $\mathcal{X}_m$ by round $s_m(i)$ in epoch $i$ is denoted by $\bar{r}_{m,n}(s_m(i))$. The sample mean reward of the empirical best arm cluster is $\bar{r}_m^*(s_m(i)) = \max_{\mathcal{K}_n \in \mathcal{A}_m(i)} \bar{r}_{m,n}(s_m(i))$.

### B. The Algorithm

Now, we describe the proposed CB-AL algorithm, whose pseudo-code is provided in Algorithm 1. Figure 2 provides an illustration of the algorithm. The algorithm operates in epochs. At the beginning of each epoch, the context/arm space partitions are determined. As aforementioned, the radii of the context and arm spaces become smaller as the epoch grows and therefore, the partitions of the spaces become finer and finer. All arm clusters for any context cluster are set to be active at the beginning of each epoch. In each time slot $t$, a context $x_t$ arrives and the algorithm finds the context cluster $\mathcal{X}_m \in \mathcal{P}_m(i)$ that it belongs to. Depending on which phase the algorithm is in (with respect to the context cluster $\mathcal{X}_m$), different operations are carried out as follows:

**Exploration**. The goal of the algorithm in the exploration phase is to explore various active arm clusters to learn their performance for $\mathcal{X}_m$. Arm clusters that are learned to be suboptimal will be de-activated over time, thereby improving the learning efficiency and system performance. In each round $s_m(i)$, the algorithm selects an active arm cluster $\mathcal{K}_n \in \mathcal{A}_m(i)$ that has not been selected in the current round for $\mathcal{X}_m$. If all active arm clusters have been selected in the current round, then the current round $s_m(i)$ ends and a new round begins. Then the algorithm arbitrarily selects any arm $k_t$ in the selected arm cluster $\mathcal{K}_n$. In the exploration phase, a query is always sent, namely $q_t = 1$, together with prior information $(a_t, b_t, \delta_t)$. The prior information is computed as follows: for round $s_m(i) > 1$, the prior information is

$$a_t = \bar{r}_{m,n} - 2L_X \rho_{X,i} - 2L_K \rho_{K,i} - 2D(s_m(i) - 1) \quad (8)$$
$$b_t = \bar{r}_{m,n} + 2L_X \rho_{X,i} + 2L_K \rho_{K,i} + 2D(s_m(i) - 1) \quad (9)$$
$$\delta_t = T_i^{-(1+\gamma)} \quad (10)$$

## Algorithm 1 Contextual Bandits with Active Learning

1: **for** epoch $i = 0, 1, 2, ...$ **do**
2:    **Initialization**: Create context and arm space partitioning $\mathcal{P}_X(i)$ and $\mathcal{P}_K(i)$. Set $\mathcal{A}_m(i) = \mathcal{P}_K(i), \forall m$. Set $\text{Stop}_m = 0, \forall m$. Set $s_m(i) = 1, \forall m$. Set $\bar{r}_{m,n} = 0, \forall m, n$.
3:    **for** time slot $t = 2^i$ to $2^{i+1} - 1$ **do**
4:       Observe the context $x_t$ and find $\mathcal{X}_m$ such that $x_t \in \mathcal{X}_m$.
5:       **switch** $\text{Stop}_m$ **do**
6:          **case** 0       ▷ **Exploration**
7:             Select $\mathcal{K}_n \in \mathcal{A}_m(i)$ that has not been selected in round $s_m(i)$ and select any $k_t \in \mathcal{K}_n$.
8:             Choose $q_t = 1$ and send the prior information $(a_t, b_t, \delta_t)$ to the annotator
9:             (A query cost $c_t$ is incurred, and the reward $\hat{r}_t = r(x_t, k_t)$ is observed.)
10:            Update $\bar{r}_{m,n}(s_m(i))$.
11:            **if** round $s_m(i)$ has finished **then**
12:               For any $\mathcal{K}_n \in \mathcal{A}_m(i)$ such that $\Delta_{m,n}(s_m(i)) \geq D_1(i, s_m(i))$, remove $\mathcal{K}_n$ from $\mathcal{A}_m(i)$.
13:               If for all $\mathcal{K}_n \in \mathcal{A}_m(i)$, $\Delta_{m,n}(s_m(i)) \leq D_2(i, s_m(i))$, then set $\text{Stop}_m = 1$.
14:               Update $s_m(i) \leftarrow s_m(i) + 1$.
15:            **end if**
16:          **case** 1       ▷ **Exploitation**
17:             Select any $\mathcal{K}_n \in \mathcal{A}_m(i)$ and any $k_t \in \mathcal{K}_n$.
18:             Choose $q_t = 0$.
19:             (The reward $r(x_t, k_t)$ is generated, but cannot be observed.)
20:    **end for**
21: **end for**

For $s_m(i) = 1$, the prior information is $a_t = 0, b_t = 1, \delta = 0$. Once the ground truth reward $r(x_t, k_t)$ is obtained from the annotator, the sample mean $\bar{r}_{m,n}$ is updated as follows

$$\bar{r}_{m,n} \leftarrow (\bar{r}_{m,n} \cdot (s_m(i) - 1) + r(x_t, k_t))/s_m(i) \quad (11)$$

At the end of a round $s_m(i)$, the algorithm de-activates suboptimal arm clusters if necessary. Specifically, the algorithm first finds the empirically best arm cluster for $\mathcal{X}_m$ and calculates the sample mean reward difference between the empirically best arm cluster and any other active arm cluster, denoted by $\Delta_{m,n}(s_m(i)) \triangleq \bar{r}_m^*(s_m(i)) - \bar{r}_{m,n}(s_m(i))$, $\forall \mathcal{K}_n \in \mathcal{A}_m(s_m(i))$. Then it compares $\Delta_{m,n}(s_m(i))$ with the current value of the control function $D_1(i, s_m(i))$. If the sample mean reward difference is greater than or equal to this value, then the corresponding arm cluster is suboptimal with high probability and hence is de-activated. Moreover, if the remaining active arm clusters have sufficiently similar sample mean reward estimates, then the algorithm stops the de-activation process in the remaining time slots of the current epoch and enters the exploitation phase. Specifically, if the reward difference for any active cluster is smaller or equal to $D_2(i, s_m(i))$, then the de-activation process stops. We denote by $S_m^i$ the number of rounds taken when the stopping condition is satisfied.

**Exploitation**. The goal of the algorithm in the exploitation phase is to exploit the best arms to maximize the reward. Since in the exploitation phase, the remaining active arm clusters are the optimal arm cluster or near-optimal arm cluster for the corresponding context cluster with high probability, the algorithm simply arbitrarily selects an active arm cluster $\mathcal{K}_n \in \mathcal{A}_m(i)$ and then arbitrarily selects an arm from $\mathcal{K}_n$. Notably, the algorithm no longer requests for the ground truth reward, i.e. $q_t = 0$, for all time slots in the exploitation phase.

## IV. REGRET ANALYSIS

To analyze the regret, we first introduce some notions.

- **Cluster reward**. We define the expected reward of selecting an arm cluster $\mathcal{K}_n$ for context cluster $\mathcal{X}_m$ as $\mu(m, n) = \max_{x \in \mathcal{X}_m, k \in \mathcal{K}_n} \mu(x, k)$. The reward of the optimal arm cluster with respect to the context cluster $\mathcal{X}_m$ is thus $\mu_m^* = \max_n \mu(m, n)$. Furthermore, we define the reward difference as $\Delta_{m,n} = \mu_m^* - \mu(m, n)$.
- **$\epsilon$-optimal arm cluster**. We define the $\epsilon$-optimal arm clusters with respect to the context cluster $\mathcal{X}_m$ as the arm clusters $\mathcal{K}_n$ that satisfy $\mu(m, n) \geq \mu_m^* - \epsilon$. Similarly, the $\epsilon$-suboptimal arm clusters are those that satisfy $\mu(m, n) < \mu_m^* - \epsilon$.
- **Normal event and abnormal event**. A normal event $\mathcal{N}_{m,n}(s_m(i))$ is an event such that the reward of selecting arm cluster $\mathcal{K}_n$ for context cluster $\mathcal{X}_m$ in round $s_m(i)$ satisfies $|\bar{r}_{m,n}(s_m(i)) - \mathbb{E}[r_{m,n}(s_m(i))]| \leq D(s_m(i))$. A abnormal event $\mathcal{N}_{m,n}^c(s_m(i))$ is an event such that $|\bar{r}_{m,n}(s_m(i)) - \mathbb{E}[r_{m,n}(s_m(i))]| > D(s_m(i))$. Further, we denote $\mathcal{N}_{i,m,n}$ as the event that no abnormal event $\mathcal{N}_{m,n}^c(s_m(i))$ occurs with respect to arm cluster $\mathcal{K}_n$ and context cluster $\mathcal{X}_m$ for the entire epoch $i$.

To analyze the regret, we first provide the following lemmas.

**Lemma 1.** *An abnormal event for arm cluster $\mathcal{K}_n$ in epoch $i$ occurs with probability at most $\delta(i) = T_i^{-\gamma}$.*

*Proof.* According to the definition of abnormal event and the Chernoff-Hoeffding bound, the probability that an abnormal

event for an arm cluster occurs in round $s_m(i)$ can be bounded by

$$\Pr\{[\mathcal{N}_{m,n}(s_m(i))]^C\} \leq 2e^{-2[D(s_m(i))]^2 s_m(i)} \leq \frac{1}{T_i^{1+\gamma}}. \quad (12)$$

Hence, the probability that an abnormal event for an arm cluster $\mathcal{K}_n$ in epoch $i$ occurs with at most

$$\sum_m \Pr\{[\mathcal{N}_{i,m,n}]^C\} \leq \sum_m \sum_{s_m(i)=1}^{S_m^i} \Pr\{[\mathcal{N}_{m,n}(s_m(i))]^C\}$$
$$\leq \sum_m \sum_{s_m(i)=1}^{S_m^i} \frac{1}{T_i^{1+\gamma}} \leq \frac{1}{T_i^{\gamma}}. \quad (13)$$

$\square$

**Lemma 2.** *(a) With probability at least $1 - N_i \delta(i)$, an $\epsilon(i)$-optimal arm clusters are not de-activated for context cluster $\mathcal{X}_m$ in epoch $i$. (b) With probability at least $1 - N_i \delta(i)$, the active set $\mathcal{A}_m(i)$ in the exploitation phase contains at most $2\epsilon(i)$-optimal arm clusters for context cluster $\mathcal{X}_m$ in epoch $i$.*

*Proof.* If the normal event occurs, for any deactivated arm clusters $\mathcal{K}_n$, we have:

$$\begin{aligned}
&\bar{r}_m^*(s_m(i)) - \bar{r}_{m,n}(s_m(i)) \\
&= (\mu_m^* - \mu(m,n)) + (\bar{r}_m^*(s_m(i)) - \mu_m^*) \\
&\quad + (\mu(m,n) - \bar{r}_{m,n}(s_m(i))) \\
&\leq \Delta_{m,n} + 2D(s_m(i)) + 2L_X \rho_{X,i} + 2L_K \rho_{K,i},
\end{aligned} \quad (14)$$

where the inequality follows from that $\bar{r}_m^*(s_m(i)) - \mu_m^* \leq D(s_m(i))$ and $\mu(m,n) - \bar{r}_{m,n}(s_m(i)) \leq D(s_m(i)) + 2L_X \rho_{X,i} + 2L_K \rho_{K,i}$. Combining with the deactivating rule, we have $\Delta_{m,n} > \epsilon(i)$.

If the normal event occurs, for any reserved active arm cluster $\mathcal{K}_n$, we also have:

$$\begin{aligned}
&\bar{r}_m^*(S_m^i) - \bar{r}_{m,n}(S_m^i) \\
&= (\mu_m^* - \mu(m,n)) + (\bar{r}_m^*(S_m^i) - \mu_m^*) \\
&\quad + (\mu(m,n) - \bar{r}_{m,n}(S_m^i)) \\
&\geq \Delta_{m,n} - 2(D(S_m^i) + L_X \rho_{X,i} + L_K \rho_{K,i}),
\end{aligned} \quad (15)$$

where the inequality follows from that $\mu_m^* - \bar{r}_m^*(S_m^i) \leq D(S_m^i) + 2L_X \rho_{X,i} + 2L_K \rho_{K,i}$ and $\bar{r}_{m,n}(S_m^i) - \mu(m,n) \leq D(s_m(i))$. Combining with the stopping rule, we have $\Delta_{m,n} \leq 2\epsilon(i)$. Since the normal event occurs with probability at least $1 - N_i \delta(i)$, the results follow. $\square$

Now we are ready to prove the regret of CB-AL.

**Theorem 1.** *The regret of the CB-AL algorithm can be upper-bounded by $R(T) = O(T^{\frac{d_X+d_K+1}{d_X+d_K+2}})$.*

*Proof.* To bound the regret, we first consider the regret caused in epoch $i$, denoted by $R_i$. This regret can be decomposed into four terms: the regret $R_i^a$ caused by abnormal events, the regret $R_i^n$ caused by $2\epsilon(i)$-optimal arm cluster selection and the inaccuracy of clusters, the regret $R_i^s$ caused by $2\epsilon(i)$-suboptimal arm cluster selection when no abnormal events occur, and the query cost $R_i^q$. We have

$$R_i \leq R_i^a + R_i^n + R_i^s + R_i^q. \quad (16)$$

Let us denote by $T_i$ the number of time slots in epoch $i$, denote by $T_{i,m}$ the number of context arrivals in context cluster $\mathcal{X}_m$ in epoch $i$, and denote by $T_{i,m,n}$ the number of query requests for arm cluster $\mathcal{K}_n$ in context cluster $\mathcal{X}_m$ in epoch $i$. We set $\alpha = \frac{1}{d_A + d_X + 2}$ and $\gamma = \frac{d_A + 1}{d_A + d_X + 2}$.

For the first term $R_i^a$ in (16), when an abnormal event happens, the regret is at most $T_i$. According to Lemma 1 abnormal events happens with probability at most $\delta(i)$ for arm cluster $\mathcal{K}_n$ in epoch $i$. Therefore, the regret $R_i^a$ in (16) can be expressed as:

$$R_i^a \leq \sum_{n=1}^{N_i} \delta(i) T_i \leq N_i \delta(i) T_i. \quad (17)$$

For the second term $R_i^n$ in (16), the regret of $2\epsilon(i)$-optimal arm cluster selection at each time slot is at most $2\epsilon(i)$, and the regret of inaccuracy of clusters at each time slot is at most $2L_X \rho_{X,i} + 2L_K \rho_{K,i}$. Therefore, the regret $R_i^n$ can be expressed as:

$$\begin{aligned}
R_i^n &\leq \sum_{t=2^i}^{2^{i+1}-1} (2\epsilon(i) + 2L_X \rho_{X,i} + 2L_K \rho_{K,i}) \\
&= 2(\epsilon(i) + L_X \rho_{X,i} + L_K \rho_{K,i}) T_i.
\end{aligned} \quad (18)$$

For the third term $R_i^s$ in (16), when the normal event occurs, according to Lemma 2, $2\epsilon(i)$-suboptimal arm cluster can only be selected in the exploration phases. Hence, the regret $R_i^s$ can be expressed as:

$$R_i^s \leq E \sum_{\Delta_{m,n} > 2\epsilon(i)} \sum_{t=2^i}^{2^{i+1}-1} \Delta_{m,n} I\{x_t \in \mathcal{X}_m, \pi_K^t \in \mathcal{K}_n, \pi_q^t = 1, \mathcal{N}_{i,m,n}\}. \quad (19)$$

According to the deactivating rule, for normal events, if the following is satisfied:

$$\Delta_{m,n} - 2D(s) - 2L_X \rho_{X,i} - 2L_K \rho_{K,i} \geq \bar{r}_m^*(s) - \bar{r}_{m,n}(s) \geq D_1(i,s), \quad (20)$$

then the arm cluster is deactivated. Hence, the rounds of exploring arm cluster $\mathcal{K}_n$, $T_{i,m,n}$ with $\Delta_{m,n} > 2\epsilon(i)$, can be bounded by

$$T_{i,m,n} \leq \frac{8 \ln(2T_i^{1+\gamma})}{[\Delta_{m,n} - (\epsilon(i) + 4L_X \rho_{X,i} + 4L_K \rho_{K,i})]^2}. \quad (21)$$

Therefore, the regret $R_i^s$ can be bounded by

$$\begin{aligned}
R_i^s &\leq E \sum_{\Delta_{m,n} > 2\epsilon(i)} \Delta_{m,n} T_{i,m,n} \\
&\leq E \sum_{\Delta_{m,n} > 2\epsilon(i)} \Big( \frac{8 \ln(2T_i^{1+\gamma})}{\Delta_{m,n} - (\epsilon(i) + 4L_X \rho_{X,i} + 4L_K \rho_{K,i})} \\
&\quad + \frac{8(\epsilon(i) + 4L_X \rho_{X,i} + 4L_K \rho_{K,i}) \ln(2T_i^{1+\gamma})}{[\Delta_{m,n} - (\epsilon(i) + 4L_X \rho_{X,i} + 4L_K \rho_{K,i})]^2} \Big) \\
&\leq \frac{8 M_i N_i \ln(2T_i^{1+\gamma})}{2\epsilon(i) - (\epsilon(i) + 4L_X \rho_{X,i} + 4L_K \rho_{K,i})} \\
&\quad + \frac{8 M_i N_i (\epsilon(i) + 4L_X \rho_{X,i} + 4L_K \rho_{K,i}) \ln(2T_i^{1+\gamma})}{[2\epsilon(i) - (\epsilon(i) + 4L_X \rho_{X,i} + 4L_K \rho_{K,i})]^2} \\
&\leq C_1 M_i N_i \ln(2T_i^{1+\gamma}) T_i^{\alpha},
\end{aligned} \quad (22)$$

where $C_1 = \frac{16L}{(L - 4L_X - 4L_K)^2}$ is a constant.

For the fourth term $R_i^q$ in (16), we first consider the query cost $R_i^{q,1}$ when the abnormal event occurs. In this case, since

the maximum query cost per slot is $2c$, the query cost can be bounded by

$$R_i^{q,1} \leq N_i 2c\delta(i)T_i. \quad (23)$$

Next, we consider the query cost $R_i^{q,2}$ in the case that only normal events occur. This can be bounded by

$$\begin{aligned}R_i^{q,2} &\leq E\sum_{m,n}\sum_{t=2^i}^{2^{i+1}-1} c_t I\{x_t \in \mathcal{X}_m, \pi_K^t \in \mathcal{K}_n, \pi_q^t = 1\}\\ &\leq E\sum_{m,n} c + E\sum_{m,n}\sum_{s=2}^{S_m^i}[c(4L_X\rho_{X,i}+4L_K\rho_{K,i}+4D(s-1))^{\beta_1}\\ &\qquad\qquad\qquad\qquad\qquad\qquad + c\eta T_i^{-(1+\gamma)\beta_2}]\\ &\leq cM_iN_i + cM_iN_i\sum_{s=2}^{S_m^i}\frac{(8L_X\rho_{X,i}+8L_K\rho_{K,i})^{\beta_1}+(8D(s-1))^{\beta_1}}{2}\\ &\qquad\qquad\qquad\qquad\qquad + c\eta M_iN_i\sum_{s=2}^{S_m^i}T_i^{-(1+\gamma)\beta_2}]\\ &\leq cM_iN_i + cM_iN_i 2^{3\beta_1-1}(L_X\rho_{X,i}+L_K\rho_{K,i})^{\beta_1}S_m^i\\ &\quad + cM_iN_i 2^{3\beta_1-1}\sum_{s=2}^{S_m^i}\frac{[\ln(2T_i^{\gamma+1})]^{\beta_1/2}}{2^{\beta_1/2}(s-1)^{\beta_1/2}} + c\eta M_iN_iT_i^{-(1+\gamma)\beta_2+1},\end{aligned}$$
(24)

where the third inequality is due to the Jensen's inequality. If $1 \leq \beta_1 < 2$, the third term on the right hand side of the last inequality in (24) can be bounded by $cN_i 2^{5\beta_1/2-1}\frac{[\ln(2T_i^{\gamma+1})]^{\beta_1/2}(S_m^i)^{1-\beta_1/2}}{1-\beta_1/2}$, due to the divergent series $\sum_{t=1}^T t^{-y} \leq T^{(1-y)}/(1-y)$ for $0 < y < 1$ [11]. If $\beta_1 \geq 2$, the third term on the right hand side of the last inequality in (24) can be bounded by $cN_i 2^{5\beta_1/2-1}[\ln(2T_i^{\gamma+1})]^{\beta_1/2}(\ln S_m^i)^{\beta_1/2}$, due to the series $\sum_{t=1}^{T-1} t^{-y} \leq \ln T$ for $y \geq 1$. We can also have the bound of $S_m^i$ (for all $m$) due to the fact that when $D_1(i, S_m^i) \leq D_2(i, S_m^i)$, the stopping rule is satisfied. Hence, $S_m^i$ can be bounded by the minimum $s$, such that $D_1(i, s) \leq D_2(i, s)$. This shows:

$$S_m^i \leq \frac{8T_i^{2\alpha}\ln(2T_i^{\gamma+1})}{(L-4L_X-4L_K)^2}. \quad (25)$$

Thus, we can bound $R_i^{q,2}$ by

$$R_i^{q,2} \leq \begin{cases} C_2 M_i N_i T_i^{\alpha(2-\beta_1)}\ln(2T_i^{\gamma+1}), & \text{if } 1 \leq \beta_1 < 2\\ C_3 M_i N_i [\ln(2T_i^{\gamma+1})]^{\beta_1}, & \text{if } \beta_1 \geq 2, \end{cases}$$
(26)

where $C_2 = c(1+\eta) + \frac{c2^{3\beta_1+5}(L_X+L_K)^{\beta_1}}{(L-4L_X-4L_K)^2} + \frac{c2^{6-\beta_1/2}}{(2-\beta_1)(L-4L_X-4L_K)^2}$ is a constant, $C_3 = c(1+\eta + 2^{5\beta_1/2-1}) + \frac{c2^{3\beta_1+5}(L_X+L_K)^{\beta_1}}{(L-4L_X-4L_A)^2}$ is a constant.

According to the definition of covering dimensions [12], the maximum number of arm clusters can be bounded by $N_i \leq C_A \rho_{K,i}^{-d_A}$ in epoch $i$, and the maximum number of context clusters can be bounded by $M_i \leq C_X \rho_{X,i}^{-d_X}$ in epoch $i$, where $C_A, C_X$ are covering constants for the arm space and the context space. Hence, the regret can be bounded by

$$\begin{aligned}R(T) &\leq E\sum_{i=0}^{\log_2 T} R^i\\ &\leq \sum_{i=0}^{\log_2 T}(\delta(i)N_i + 2c\delta(i)N_i + 2\epsilon(i) + 2L_X\rho_{X,i} + 2L_K\rho_{K,i})T_i\\ &\quad + E\sum_{i=0}^{\log_2 T} O(1)\ln(2T_i^{1+\gamma})T_i^\alpha\\ &\leq O(1)T^{\frac{d_X+d_A+1}{d_X+d_A+2}}\ln(T).\end{aligned}$$
(27)

Therefore, the result of Theorem 1 follows. □

We further show a lower bound for the CB-AL algorithm. Since the proposed algorithm incurs the query cost when it requests a ground truth, the lower bound of the regret cannot be lower than that of the conventional contextual MAB setting where no query cost is incurred [1].

**Theorem 2.** *The regret of the CB-AL algorithm can be lower-bounded by $R(T) = \Omega(T^{\frac{d_X+d_K+1}{d_X+d_K+2}})$.*

Theorem 1 and Theorem 2 together show that our algorithm is order-optimal and achieves the same order as conventional contextual bandits algorithms in cost-free scenarios [1][2][3].

## V. NUMERICAL RESULTS

We conduct illustrative experiments using synthetic data with 2-dimensional contexts and 2-dimensional arms. In our first experiment, we compare the performance of our proposed CB-AL algorithm with the Contextual Bandit algorithm and the Contextual Bandit Active Learning without considering prior information (CB-AL without prior information). The result is shown in Fig. 3. As we can see, the proposed CB-AL algorithm performs better, in terms of payoff, than the conventional bandit algorithm (by $16\%$) and the CB-AL without prior information (by $13\%$) by the end of the experiment duration.

In our second experiment, we show the payoffs achieved by our proposed CB-AL algorithm when the query cost varies (by changing cost parameters $c$ in e.q. (1)). We show the result in Fig 4. We can see that as $c$ increases from 0.1 to 1, the achieved payoff decreases by $15\%$ for $T = 10000$ and by $8\%$ for $T = 20000$.

## VI. CONCLUSIONS

In this paper, we developed a contextual bandits learning algorithm with active learning capability. The active learning cost is reduced by providing prior information about the reward realization to the annotator. The algorithm maintains and updates partitions of the context and arm spaces, and operates between exploration and exploitation phases. Through precise control of the partitioning process and when to request the ground truth of the reward, the algorithm gracefully balances the accuracy of learning and the cost incurred by active learning. We prove that the regret of the proposed algorithm achieves the same order as that of conventional contextual bandits algorithms in cost-free scenarios.

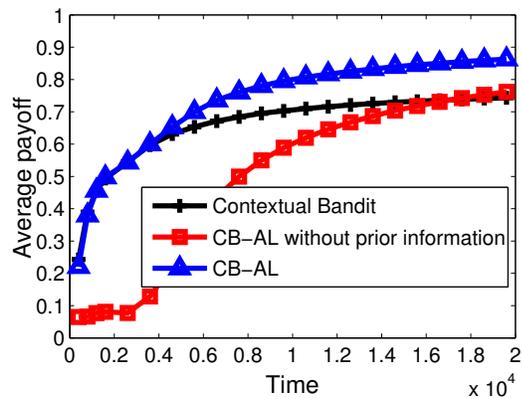

Fig. 3. Comparison of performance for different algorithms.

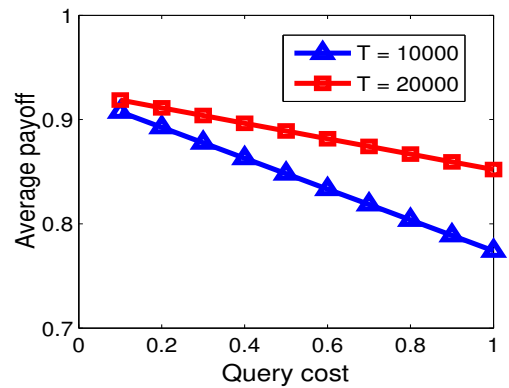

Fig. 4. Relationship between payoff and cost.

## REFERENCES

[1] T. Lu, D. Pál, and M. Pál, "Contextual multi-armed bandits," in *Artificial Intelligence and Statistics Conference (AISTATS)*, 2010, pp. 485–492.
[2] J. Langford and T. Zhang, "The epoch-greedy algorithm for multi-armed bandits with side information," in *Advances in Neural Information Processing Systems*, 2008, pp. 817–824.
[3] A. Slivkins, "Contextual bandits with similarity information." *Journal of Machine Learning Research*, vol. 15, no. 1, pp. 2533–2568, 2014.
[4] L. Song, W. Hsu, J. Xu, and M. van der Schaar, "Using contextual learning to improve diagnostic accuracy: Application in breast cancer screening," *IEEE Journal of Biomedical and Health Informatics*, vol. 20, no. 3, pp. 902–914, 2016.
[5] J. Xu, T. Xing, and M. van der Schaar, "Personalized course sequence recommendations," *IEEE Transactions on Signal Processing*, vol. 64, no. 20, pp. 5340–5352, 2016.
[6] B. Settles, "Active learning literature survey," *University of Wisconsin, Madison*, vol. 52, no. 55-66, p. 11, 2010.
[7] D. A. Cohn, Z. Ghahramani, and M. I. Jordan, "Active learning with statistical models," *Journal of Artificial Intelligence Research*, vol. 4, no. 1, pp. 129–145, 1996.
[8] M.-F. F. Balcan and V. Feldman, "Statistical active learning algorithms," in *Advances in Neural Information Processing Systems*, 2013, pp. 1295–1303.
[9] A. K. McCallumzy and K. Nigamy, "Employing em and pool-based active learning for text classification," in *Proc. International Conference on Machine Learning (ICML)*, 1998, pp. 359–367.
[10] S. Dasgupta, "Analysis of a greedy active learning strategy." in *Advances in Neural Information Processing Systems*, vol. 17, 2004, pp. 337–344.
[11] E. Chlebus, "An approximate formula for a partial sum of the divergent p-series," *Applied Mathematics Letters*, vol. 22, no. 5, pp. 732–737, 2009.
[12] J. Heinonen, *Lectures on analysis on metric spaces*. Springer Science & Business Media, 2012.